\documentclass{amia}
\usepackage{lipsum} 
\usepackage{multirow}
\usepackage{amssymb}
\usepackage{amsmath}
\usepackage{booktabs}
\usepackage{color}
\usepackage{algorithm, algpseudocode}
\usepackage{amssymb}
\usepackage{pifont}
\newcommand{\cmark}{\ding{51}}
\newcommand{\xmark}{\ding{55}}
\usepackage{makecell}

\def\Algnameabbr{FairPM}

\begin{document}

\title{Towards Fair Patient-Trial Matching via Patient-Criterion Level Fairness Constraint}

\author{Chia-Yuan Chang$^1$, Jiayi Yuan$^2$, Sirui Ding$^1$, Qiaoyu Tan$^1$, Kai Zhang, PhD$^3$, \\ Xiaoqian Jiang, PhD$^3$, Xia Hu, PhD$^2$, Na Zou, PhD$^1$ }

\institutes{
    $^1$ Texas A\&M University, College Station, TX, USA; $^2$Rice University, Houston, TX, USA; $^3$University of Texas Health Science Center, Houston, TX, USA.
}

\maketitle

\section*{Abstract}
Clinical trials are indispensable in developing new treatments, but they face obstacles in patient recruitment and retention, hindering the enrollment of necessary participants. To tackle these challenges, deep learning frameworks have been created to match patients to trials. These frameworks calculate the similarity between patients and clinical trial eligibility criteria, considering the discrepancy between inclusion and exclusion criteria. Recent studies have shown that these frameworks outperform earlier approaches. However, deep learning models may raise fairness issues in patient-trial matching when certain sensitive groups of individuals are underrepresented in clinical trials, leading to incomplete or inaccurate data and potential harm.
To tackle the issue of fairness, this work proposes a fair patient-trial matching framework by generating a patient-criterion level fairness constraint. The proposed framework considers the inconsistency between the embedding of inclusion and exclusion criteria among patients of different sensitive groups. The experimental results on real-world patient-trial and patient-criterion matching tasks demonstrate that the proposed framework can successfully alleviate the predictions that tend to be biased.

\section{Introduction}


Clinical trials are an essential part of developing new treatments for diseases, as they provide rigorous scientific evidence for the safety and efficacy of new therapies. Despite their importance, clinical trials often face significant challenges in patient recruitment and retention due to the difficulty in obtaining the required number of participants. 
Several studies have examined the factors that affect patient participation in clinical trials. For instance, an existing study found that inadequate patient participation and recruitment can lead to delays and increased costs in clinical trials, ultimately hindering the development of new treatments~\cite{ohmann2017sharing}. A recent study also found that nearly one-third of publicly funded trials required time extensions due to low enrollment rates~\cite{campbell2007recruitment}.


Recently, patient-trial matching has been the focus of research to accurately identify and recruit qualified patients. The existing patient-trial matching methods could be divided into two categories: rule-based systems and machine learning approaches, which both have been proposed to accelerate the patient recruiting process.
Rule-based systems rely on a vast number of human annotations to establish classification rules~\cite{weng2011elixr, kang2017eliie}. However, they have limitations in terms of recall because of inadequate rule coverage, and require extensive manual efforts to set up rules. Alternatively, machine learning based models focus on extracting rules automatically. For instance, \cite{alicante2016unsupervised} adopts unsupervised clustering methods to automatically extract eligible rules.
More recently, several studies also employ deep neural networks to further improve the model performance on patient-trial matching.
For example, DeepEnroll~\cite{zhang2020deepenroll} and COMPOSE~\cite{gao2020compose} proposed utilizing deep embedding models to encode patient records and eligibility criteria of clinical trials for computing the similarity between patients and criteria in the embedding space. By considering the discrepancy between inclusion and exclusion criteria in clinical trials, these frameworks have shown promise in achieving more precise predictions and efficient matching.

However, there are disparity issues in patient-trial matching, which can be further amplified by machine learning models and lead to unfairness. Specifically, machine learning models trained on biased historical data can perpetuate disparities in patient-trial matching, resulting in underrepresented sensitive groups of individuals in clinical trials and limited treatment efficacy.
Unfortunately, recent research has shed light on the potential for machine learning models to exhibit unfairness and bias~\cite{ding2023fairly}, which may negatively impact the minority groups in the application fields. For example, a study found that a machine learning algorithm used to predict healthcare utilization showed bias against African-American patients, resulting in fewer healthcare resources allocated to them compared to white patients~\cite{obermeyer2019dissecting}. Current studies concentrate on developing bias mitigation methods to reduce discrimination in machine learning models.
There are several existing fairness regularization~\cite{beutel2019putting, beutel2019fairness, jiang2020wasserstein, nam2020learning} and adversarial learning methods~\cite{zhang2018mitigating, madras2018learning, sweeney2020reducing} are designed to ensure fairness by preventing discrimination based on sensitive attributes such as race, gender, or age.
However, the existing fairness methods cannot be utilized to mitigate the fairness issue in patient-trial matching because of the complex inclusion and exclusion criteria, which require careful consideration to achieve accurate prediction.

The uniqueness of the patient-trial matching lies in its dual goals of matching inclusion criteria while mismatching exclusion criteria, which differs from other healthcare applications and adds extra complexity to the task.
Specifically, the discrepancy between inclusion and exclusion criteria provides information about a clinical trial to better learn a patient-trial matching framework. 
To better characterize the uniqueness and tackle the fairness challenges, we propose \Algnameabbr{}, a fine-grained fairness framework for patient-trial matching tasks.
Specifically, motivated by DeepEnroll~\cite{zhang2020deepenroll} and COMPOSE~\cite{gao2020compose}, we develop a patient-trial matching framework by minimizing the distance between the embedding of qualified patients and inclusion criteria while maximizing the distance between the embedding of unqualified patients and exclusion criteria. To further mitigate the biased prediction behaviors, we propose a fine-grained fairness constraint to minimize the prediction differences among the inclusion and exclusion criteria and across different sensitive patient groups.
We evaluate the proposed framework on a real-world EHR patient records dataset and six pivotal stroke clinical trials. The experimental results demonstrate that \Algnameabbr{} can improve two fairness metrics for both patient-criterion and patient-trial matching toward two sensitive attributes, albeit with a slight trade-off in prediction performance.
The case study shows some eligibility criteria that may cause biased predictions for minority groups.


\section{Background of Fairness in Patient-Trial Matching}

In this section, we will first identify the fairness issue in the patient-trial matching from two levels, and then introduce the metrics to measure them from the computational perspective.

\subsection{Fairness of patient-trial matching}
We identified two critical fairness issues in matching patients with trials in previous matching systems, namely \emph{criteria-level} and \emph{trial-level} fairness. Criteria-level fairness indicates that criteria assessment should be consistent across all patient subgroups. For example, clinical trial eligibility criteria should be assessed the same way for patients of the majority and minority races. Conversely, trial-level fairness mean that different patient subgroups for the same clinical trial should be equal considered. For example, male and female patients in a clinical trial should have equal eligibility that is unrelated to gender. Although abundant of machine learning efforts have been made to predict patient eligibility for different clinical trials, they often ignore the fairness issues behind the clinical matching, as discussed before. Therefore, there is an urgent need to enable ML models product unbiased eligibility predictions from both the criteria and trial perspectives.

\subsection{Fairness metrics}

Fairness metrics have garnered considerable attention in recent years. For example, research in \cite{binns2018fairness} delved into fairness definitions within political philosophy, attempting to establish connections with machine learning principles. Another study in \cite{hutchinson201950} examined the evolution of fairness definitions over a period of five decades, focusing on the fields of education and machine learning. Additionally, comprehensive investigations have been conducted to enumerate and elucidate various definitions of fairness as they pertain to algorithmic classification challenges (\cite{verma2018fairness, mehrabi2021survey}). In the subsequent section, we will reiterate and expound upon some of the most widely adopted definitions in our work.


\textbf{Equal Opportunity (EO).} It is a binary predictor $\hat{Y}$ that adheres to the principle of equal opportunity with respect to protected attribute $\hat{A}$ and outcome $\hat{Y}$, if $P(Y=1|A=0,Y=1) = P(\hat{Y}=1|A=1,Y=1)$ \cite{hardt2016equality}. This assertion implies that the likelihood of an individual belonging to the positive class being allocated a positive outcome should be equivalent for both protected and unprotected group members \cite{verma2018fairness}. Thus, the equal opportunity definition stipulates that the true positive rates should be consistent across both protected and unprotected groups.

\textbf{Demographic Parity (DP).} Alternatively referred to as statistical parity, it is a predictor $Y$ upholds demographic parity if $P(Y|A=0) = P(\hat{Y}|A=1)$ \cite{dwork2012fairness, kusner2017counterfactual}. This principle dictates that the probability of a positive outcome \cite{verma2018fairness} should remain consistent irrespective of an individual's membership in the protected group. In other words, demographic parity mandates that the likelihood of a positive outcome should be independent of the protected attribute.

\section{Data and Problem Description}

\subsection{Data preparation}

In the present investigation, we have undertaken a comprehensive analysis of data obtained from the renowned Texas Medical Center. This research focuses on six pivotal stroke trials, i.e., NCT03735979, NCT03805308, NCT03263117, NCT03496883, NCT03876457, and NCT03545607. 
The patient data are participants encompassed in these studies amounts to 825 individuals. The project was approved by the UTHealth Institutional Review Board (IRB) under HSC-SBMI-21-0529 - ``Re-admission Risk Estimation for Stroke Patients''.
For the purpose of this study, race and gender have been identified as sensitive demographic groups, warranting further examination. A meticulous breakdown of the demographic characteristics for these patients can be found in Table~\ref{tab:demo}.

\vspace{0.5em}
\begin{table}[ht]
  \centering
  \caption{The demographic information pertaining to gender and race within the dataset.}
\begin{tabular}{|cccc|}
\hline
\textbf{Dataset} & \textbf{Male\ /\ Female} & \textbf{White\ /\ Others} & \textbf{Total} \\ \hline
Train & 308\ /\ 217 & 185\ /\ 340 & 515 \\ \hline
Valid & 28\ /\ 31 & 18\ /\ 41 & 59 \\ \hline
Test & 135\ /\ 116 & 83\ /\ 168 & 251 \\ \hline
\end{tabular}
  \label{tab:demo}%
\end{table}%

\subsection{Problem formulation}
\label{sec:prob_def}
In this section, we will go over the notations and formulate the problems in this paper. We first define the notations for describing patient records and then introduce the two main tasks in this paper.

\noindent\textbf{Definition 1: Patient records.} We use $P = [v_1, v_2, \cdots, v_T]$ to represent a series of patient visit records 
within the longitudinal electronic health records (EHR). Every visit record contains three groups of observations: diagnosis $\mathcal{D}$, medication $\mathcal{M}$, and procedure $\mathcal{P}$. These groups correspond to sets of diseases, medication types, and procedural operations, respectively. Given the three observation groups, each visit record of a patient can be represented by $v_t = [d_{t_1}, d_{t_2}, \cdots, d_{t_i}, m_{t_1}, m_{t_2}, \cdots, m_{t_j}, p_{t_1}, p_{t_2}, \cdots, p_{t_k}]$, where $d_{t_i} \in \mathcal{D}$, $m_{t_i} \in \mathcal{M}$, and $p_{t_i} \in \mathcal{P}$. Since all the medical codes in $\mathcal{D}$, $\mathcal{M}$, and $\mathcal{P}$ are frequently utilized and can be considered as a single general concept, we represent them as $g_t$ for the sake of simplicity.

In this work, the sensitive attribute for each patient visit record is defined as $v_s \in \boldsymbol{S}$, where $\boldsymbol{S}$ is the sensitive attributes set, and the target sensitive groups place particular emphasis on the attributes of race and gender.

\noindent\textbf{Definition 2: Clinical trials.} Each clinical trial consists of two categories of eligibility criteria: inclusion criteria ($c^{I}$) and exclusion criteria ($c^{E}$). Therefore, we can denote each clinical trial as $C = [c^{I}_1, c^{I}_2, \cdots, c^{I}_N, c^{E}_1, c^{E}_2, \cdots, c^{E}_Q]$, where $N$ and $Q$ represent the number of inclusion and exclusion criteria, correspondingly. Note that each criterion is described in text.

\noindent\textbf{Task 1: Patient-Criterion matching.} When given the visit records of a patient $P$ and an inclusion or exclusion criterion belonging to a clinical trial, we define the matching of patient-trial as a multi-class classification task. It means that a pair of patient $P$ and criterion $c$ can be classified into three possible categories: "inclusion," "exclusion," and "unknown." These three categories show whether the criteria include or exclude the patient. We can represent the patient-criterion matching task as $\hat{y}(P, c) \in \{inclusion, exclusion, unknown\}$.

\noindent\textbf{Task 2: Patient-Trial matching.} When given the visit records of a patient $P$ and a clinical trial $C$, we define the patient-trial matching as a binary classification task, indicating whether a patient $P$ is eligible for the clinical trial $C$. For a patient $P$ to be eligible for the clinical trial $C$, all of the inclusion criteria $c^I \in C$ must apply to the patient and none of the exclusion criterion $c^E \in C$ should apply to the patient. In other words, the patient-trial matching task is a 100\% patient-criterion matching task where all the criteria belong to the same clinical trial $C$.

\section{Fair Patient-Trial Matching (\Algnameabbr{})}

\begin{figure*}[t!]
\centerline{\includegraphics[width=0.9\textwidth]{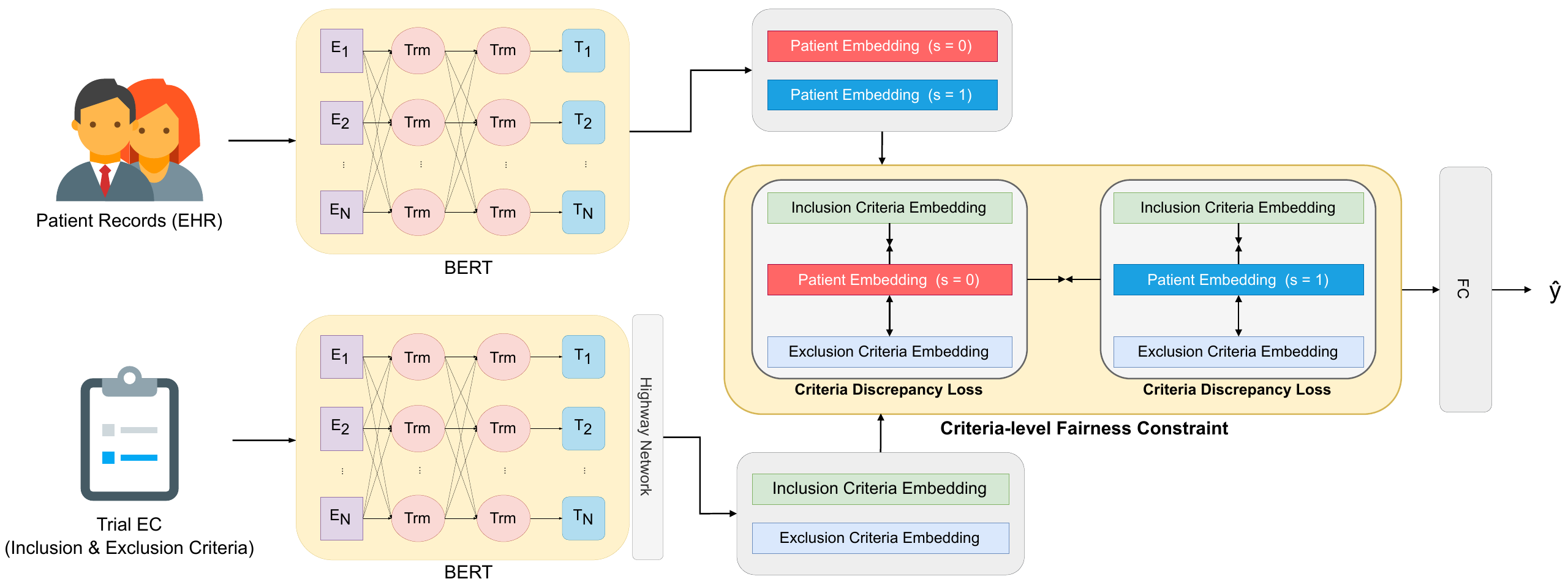}}
\caption{
An overview of the proposed framework \Algnameabbr{}, where BERTs refer to the encoding models for patient records and eligibility criteria (see Section~\ref{sec:text_encoder}), Criteria Discrepancy Loss represents the goal of optimizing the distance between patients and criteria (see Section~\ref{sec:task_loss}), and Criteria-level Fairness Constraint refers to the fairness constraint considering the discrepancy among the eligibility criteria and sensitive groups (see Section~\ref{sec:fair_loss}).}
\label{fig:model_layout}
\end{figure*}

Inspired by DeepEnroll~\cite{zhang2020deepenroll} and COMPOSE~\cite{gao2020compose}, we propose the \Algnameabbr{} framework that leverages deep embedding models for mapping input data to latent space and predict whether a patient is eligible for a criteria. To further tackle fairness issue, we introduce a specific fairness constraint that is tailored to the characteristics of patient-criterion and patient-trial matching tasks.

In this section, we present our \Algnameabbr{} to achieving fair patient-criterion and patient-trial matching. Figure~\ref{fig:model_layout} depicts an overview of the proposed \Algnameabbr{}, which comprises text encoder, criteria discrepancy loss, fair criteria matching loss, and a fully connected layer serving as the prediction head.
First, we introduce the encoding model of patient records and eligibility criteria into embedding space (Section~\ref{sec:text_encoder}). Next, we present a joint learning approach to distinguish between inclusion and exclusion criteria for patient-criteria and patient-trial matching (Section~\ref{sec:task_loss}). We then propose a fine-grained, criteria-level fairness constraint to achieve fair patient-trial matching (Section~\ref{sec:fair_loss}). Finally, we describe the training algorithm in its entirety (Section~\ref{sec:algo}).

\subsection{Embedding of Patient Records and Eligibility Criteria}
\label{sec:text_encoder}
\noindent\textbf{Patient Records Embedding.} In a single patient visit record, the diagnosis results can include information on diagnosed diseases, medications, and procedures, all of which are expressed in natural language. Hence, latent representations of a patient's visit record can be learned to use large language models (LLMs), such as BERT~\cite{devlin2018bert} and RoBERTa~\cite{liu2019roberta}. In this work, we leverage the pretrained BERT as the text encoder.
Furthermore, since a patient is represented by a sequence of past diagnosis records, we employ a memory network~\cite{weston2014memory} to process the patient records, which helps to effectively preserve the sequence of visit information in the embedding space. 
Formally, a patient record embedding, denoted by $z_P$, is obtained through the encoding process $g_P(\cdot)$, which can be represented as follows:
\begin{align}
    z_P 
    \notag &= g_P(P) \\
    \notag &= M_{mem}(BERT(v_1), BERT(v_2), \cdots, BERT(v_T)) \\
    &= M_{mem}(\boldsymbol{d}, \boldsymbol{m}, \boldsymbol{p}),
\label{eq:patient_encoder}
\end{align}
where $\boldsymbol{d}$, $\boldsymbol{m}$, and $\boldsymbol{p}$ denote the aggregated historical diagnosis, medications, and procedures embeddings, respectively. We incorporate a memory network $M_{mem}(\cdot)$ to preserve the sequence of visits information in the embedding space.

\noindent\textbf{Eligibility Criteria Embedding.} Each clinical trial is described by its eligibility criteria (ECs), including both inclusion and exclusion criteria, in unstructured textural description. Therefore, the embedding of each EC can also be learned by LLMs, and we use BERT, the same as for patient record embedding. However, one salient characteristic of ECs is the frequent appearance of concepts that express significant and detailed information, including numerical values and associated quantity units. To capture and encode these crucial features in the embedding space, we adopt a previous approach of using a convolutional neural network (CNN)~\cite{hu2014convolutional} and a highway layer~\cite{srivastava2015highway} to extract patterns across multiple levels for the semantic matching task~\cite{you2018end}. Formally, the encoding process $g_c(\cdot)$ is used to encode an EC embedding $z_c$, which can be formulated as follows:
\begin{align}
    z_c 
    \notag &= g_c(c) \\
    \notag &= Highway(Conv(BERT(c))) \\
    &= \sigma(Conv(BERT(c))) \cdot Conv(BERT(c)) + Conv(BERT(c)) \cdot (1 - \sigma(Conv(BERT(c)))),
\label{eq:ec_encoder}
\end{align}
where, $\sigma(\cdot)$ represents the sigmoid activation function. Generally, we utilize the Highway layer $Highway(\cdot)$ to effectively capture the specific semantic concepts present in the ECs.

\subsection{Joint Criteria Discrepancy Loss}
\label{sec:task_loss}
Since the patient-trial matching task depends on the prediction results of patient-criteria matching, the objective of our learning workflow is to optimize the patient-criteria matching task. However, a crucial characteristic of the patient-criteria matching task is the discrepancy between the patient-inclusion and patient-exclusion criteria pairs. To train the framework taking the uniqueness of the task into account, we follow the objective loss proposed by the framework COMPOSE~\cite{gao2020compose}. Specifically, the objective loss includes two parts as the following.

\noindent\textbf{Cross-entropy Loss.} Since the patient-criteria matching is formulated as a multi-class classification problem (Section.~\ref{sec:prob_def}), we can develop a classification framework by optimizing a cross-entropy loss between the predicted labels $\hat{y}$ and the ground truth $y$ as follows:
\begin{align}
    \mathcal{L}_{CE} = -(y^{T} \cdot log(\hat{y}) + (1 - y)^{T} \cdot log(1 - \hat{y}))
\label{eq:cross_entropy}
\end{align}

\noindent\textbf{Criteria Discrepancy Loss.} Considering the characteristic of the patient-criteria matching task that the inclusion and exclusion criteria can have opposite effects, we adopt a loss that accounts for the difference between them. Specifically, the goal of the criteria discrepancy loss is to minimize the distance between the embedding of qualified patients and the inclusion criteria $c^I$, while maximizing the distance between the embedding of unqualified patients and the exclusion criteria $c^E$. Formally, the criteria discrepancy loss can be formulated as follows:
\begin{equation}
    \mathcal{L}_{CD} =
    \left \{
    \begin{aligned}
        & 1 - d(z_P, z^{I}_c), & if\: z_P\: is\: z_{P, c^{I}} \\
        & max(0, d(z_P, z^{E}_c) - \kappa), & if\: z_P\: is\: z_{P, c^{E}} \\
    \end{aligned}
    \right.
\label{eq:criteria_dis_loss}
\end{equation}
where $d(\cdot, \cdot)$ is an arbitrary distance function in metric space, $z_{P, c^I}$ represents the embedding of the patient who matches the inclusion criteria $c^I$, $z_{P, c^E}$ represents the embedding of the patient who doesn't match the exclusion criteria $c^E$, and $\kappa$ denotes the hyper-parameter of minimum distance between $z_P$ and $z^{E}_c$.

Combination of the cross-entropy and criteria discrepancy loss, the joint objective loss for learning a patient-criteria matching framework is as follows:
\begin{align}
    \mathcal{L}_{PC} = \mathcal{L}_{CE} + \mathcal{L}_{CD}.
\label{eq:obj_loss}
\end{align}

\subsection{Criteria-level Fairness Constraint}
\label{sec:fair_loss}
Despite the prediction efficacy of the LLM encoders and the designed objective function, the skewed demographic distribution in the training data inherently causes fairness issues for the trained deep models. However, directly adopting existing debiasing regularization without considering the unique characteristics of the training target may lead to the exacerbation of unfair predictions against protected groups. Therefore, it is necessary to carefully tailor the bias mitigating approach to the specific matching task.

To address the potential fairness issue resulting from the discrepancy between patient-inclusion and patient-exclusion criteria matching, we propose a novel approach that aims to minimize the prediction differences among the two types of criteria and across different sensitive patient groups. By doing so, we can mitigate the potential impact of biased model predictions on certain subgroups of patients. Formally, the proposed criteria-level fairness constraint can be formulated as follows:
\begin{align}
    \mathcal{L}_{FC} = \sum_{i \in \textbf{S}} \sum_{j \in \textbf{S}/i} |\mathcal{L}_{CD, [v_s = i]} - \mathcal{L}_{CD, [v_s = j]}|,
\label{eq:fair_loss}
\end{align}
where $\mathcal{L}_{CD, [v_s = i]}$ represents the criteria discrepancy loss $\mathcal{L}_{CD}$ of a patient who belongs to the sensitive group $i$.

Finally, we can learn a fair patient-criteria and patient-trial matching framework by optimizing the joint criteria discrepancy loss $\mathcal{L}_{PC}$ with the criteria-level fairness constraint $\mathcal{L}_{FC}$:
\begin{align}
    \mathcal{L} = \mathcal{L}_{PC} + \lambda_{FC} \mathcal{L}_{FC},
\label{eq:total_loss}
\end{align}
where $\lambda_{FC}$ is the weighting hyper-parameter to balance the fairness constraint and the performance of predictions.

\subsection{Algorithm of \Algnameabbr{} Training}
\label{sec:algo}
The training outline of the proposed \Algnameabbr{} framework is given in Algorithm~\ref{alg:training}. The training aims to achieve the fair patient-criteria and patient-trial matching framework by optimizing the joint objective loss $\mathcal{L}_{PC}$ with the proposed task-specific fairness constraint $\mathcal{L}_{FC}$. Specifically, \Algnameabbr{} first encode the patient records and eligibility criteria to embedding space (line 4-5), and then update the patient records encoder, eligibility criteria encoder, and predictor according to Eq.~\ref{eq:obj_loss} and Eq.~\ref{eq:fair_loss} (line 6) until it converges.

\begin{algorithm}
    \caption{Algorithm of Fair Patient-Criteria Matching (\Algnameabbr{}) Training}
    \label{alg:training}
    \begin{algorithmic}[1]
      \State {\bfseries Input:} 
      
      A set of patient records comprises multiple visit records $\boldsymbol{P}$

      A set of eligibility criteria including inclusion criteria $\boldsymbol{c^{I}}$ and exclusion criteria $\boldsymbol{c^{E}}$
      
      Patient records encoder $g_P(\boldsymbol{\cdot})$

      Eligibility criteria encoder $g_c(\boldsymbol{\cdot})$
      
      Fully connection predictor $F(\cdot)$
      
      \State {\bfseries Output:} 
      
      Fair patient records encoder $g_P(\boldsymbol{P})$ eligibility criteria encoder $g_c(\boldsymbol{c})$, and fully connection predictor $F(\cdot)$
      \While{not convergence}
      \State Encode a set of patient records to embedding space $g_P(\boldsymbol{P}) = \boldsymbol{z_P}$
      \State Encode eligibility criteria of a set of clinical trials to embedding space $g_c(\boldsymbol{c}) = \boldsymbol{z_c}$
      \State Update $g_P(\cdot)$, $g_c(\cdot)$, and $F(\cdot)$ by optimizing the task loss Eq.~\ref{eq:obj_loss} with the fairness constraint Eq.~\ref{eq:fair_loss}
      \EndWhile
    \end{algorithmic}
\end{algorithm}

\section{Experiment}
\label{sec:exp}

In this section, we conduct experiments to evaluate the performance of \Algnameabbr{} framework, aiming to answer the following three research questions:

\begin{itemize}
    \item \textbf{RQ1:} How effective is the \Algnameabbr{} for improving different sensitive attributes (Section~\ref{sec:exp_results})?
    \item \textbf{RQ2:} How does the hyper-parameter $\lambda_{FC}$ impact the fairness performance of \Algnameabbr{} (Section~\ref{sec:sensitive_hyper})?
    \item \textbf{RQ3:} What is the difference between the results of \Algnameabbr{} and the vanilla model (Section~\ref{sec:case_study})?
\end{itemize}

\subsection{Baseline methods}
\vspace{-0.1cm}
\textbf{Baseline Model.} 
COMPOSE~\cite{gao2020compose} outperformed other baseline models, including LSTM+GloVE~\cite{hochreiter1997long}, LSTM+BERT~\cite{devlin2018bert}, Criteria2Query~\cite{yuan2019criteria2query}, and DeepEnroll~\cite{zhang2020deepenroll}. As our \Algnameabbr{} framework is inspired by and developed based on COMPOSE, we refer to the version of \Algnameabbr{} removing the proposed task-specific fairness constraint $\mathcal{L}_{FC}$ (Eq.~\ref{eq:fair_loss}) as the baseline model for simplification.

\textbf{Baseline Fairness Adversarial Learning Constraint (Baseline w/ ALC).} In the seminal work of 2018, Zhang et al. introduced a novel approach utilizing adversarial networks as a means of mitigating model bias \cite{zhang2018mitigating}. This innovative methodology was derived from the concept of generative adversarial networks \cite{goodfellow2020generative}. The framework devised by Zhang et al. involved training the generator with a specific focus on a protected attribute, such as gender, ultimately leading to a structure in which the generator actively obstructs the discriminator's ability to predict gender within a given overarching task. The proposed adversarial learning to mitigate bias successfully demonstrated an enhancement in the fairness of an income classification task. However, this improvement was accompanied by a slight reduction in overall accuracy. We implement this adversarial debiasing method as our baseline fairness constraint, named ALC.

\subsection{Evaluation tasks and metrics}
\vspace{-0.1cm}
As mentioned in Section~\ref{sec:prob_def}, there are two evaluation tasks: 

\noindent\textbf{Patient-Criterion matching.} We begin by labeling all patient-criterion pairs, and then splitting them into training, validation, and testing sets. The training set is used for model training, the validation set for hyperparameter tuning, and the testing set for evaluating both the baselines and the proposed \Algnameabbr{}.

\noindent\textbf{Patient-Trial matching}: We define the patient-trial matching task as a binary classification problem that determines whether a patient $P$ is qualified for a clinical trial $C$. A patient $P$ is considered eligible for a clinical trial $C$ only if they satisfy all the inclusion criteria $c^I$ and do not satisfy any of the exclusion criteria $c^E$ in the clinical trial $C$. The training, validation, and testing sets are split in the same way as for the patient-criterion matching task.

For both tasks, we evaluate the prediction performance using accuracy score (\textbf{Acc.}) and F1 score (\textbf{F1}), and assess the fairness of the models using demographic parity (\textbf{DP}) and equal opportunity (\textbf{EO}) as evaluation metrics.

\subsection{Implementation details}
\vspace{-0.1cm}
We implement all the baseline models and \Algnameabbr{} in PyTorch. As for the LLM text encoders, we leverage the Clinical BERT~\cite{alsentzer2019publicly} pretrained on PubMed and MIMIC-III~\cite{johnson2016mimic} for increasing the medical knowledge of the text embedding.

\subsection{Prediction performance}
\label{sec:exp_results}
\vspace{-0.1cm}
We conduct experiments to compare the prediction and fairness performance of the proposed \Algnameabbr{} with other baselines on patient-criterion and patient-trial matching tasks. Our target sensitive attributes are \textit{Race} and \textit{Gender}.

\begin{table*}[ht]
\renewcommand{\arraystretch}{1.2}
\centering
\vspace{0.2 cm}
\caption{Performance comparison on patient-criteria matching task}
\vspace{-0.2 cm}
\begin{tabular}{|c|cccc|cccc|}
\hline
\multirow{2}{*}{\textbf{Model}} & \multicolumn{4}{c|}{\textbf{Sensitive attribute: Race}}         & \multicolumn{4}{c|}{\textbf{Sensitive attribute: Gender}}         \\ \cline{2-9} 
                                 & \textbf{Acc.} & \textbf{F1} & \textbf{DP} & \textbf{EO} & \textbf{Acc.} & \textbf{F1} & \textbf{DP} & \textbf{EO} \\ \hline
Baseline                         & 0.9595   & 0.9702    & 0.0301    & 0.0251    & 0.9595    & 0.9702    & 0.0216    & 0.0093     \\ \cline{1-9} 
Baseline w/ ALC                 & 0.8294       & 0.8839        & 0.0112        & 0.0248        & 0.8142        & 0.8642        & 0.0095        & 0.0010        \\ \cline{1-9} 
\Algnameabbr{}                   & 0.9130   & 0.9360    & 0.0001    & 0.0111    & 0.9299    & 0.9490    & 0.0093    & 0.0011    \\ \cline{1-9} 
\end{tabular}
\label{tab:exp_criteria}
\end{table*}

\noindent\textbf{Results of patient-criterion matching.} 
Table~\ref{tab:exp_criteria} summarizes the performance of \Algnameabbr{} and the two baseline methods on the testing set. Since the baseline model is only trained by the joint objective including two prediction task-oriented loss terms, cross-entropy $\mathcal{L}_{CE}$ and criteria discrepancy loss $\mathcal{L}_{CD}$, it can achieve the best accuracy and F1 scores.
However, the baseline model may learn the skew distribution of sensitive attributes in training data, which can be addressed by employing ALC~\cite{goodfellow2020generative} to the baseline model. Nevertheless, as shown in Table~\ref{tab:exp_criteria}, adopting ALC would cause a huge performance drop.
Comparing with the two baselines, \Algnameabbr{} improve the fairness metrics DP and EO, while maintaining the competitive performance for the patient-criterion matching task.

\begin{table*}[ht]
\renewcommand{\arraystretch}{1.2}
\centering
\vspace{0.2 cm}
\caption{Performance comparison on patient-trial matching task}
\vspace{-0.2 cm}
\begin{tabular}{|c|cccc|cccc|}
\hline
\multirow{2}{*}{\textbf{Model}} & \multicolumn{4}{c|}{\textbf{Sensitive attribute: Race}}         & \multicolumn{4}{c|}{\textbf{Sensitive attribute: Gender}}         \\ \cline{2-9} 
                                 & \textbf{Acc.} & \textbf{F1} & \textbf{DP} & \textbf{EO} & \textbf{Acc.} & \textbf{F1} & \textbf{DP} & \textbf{EO} \\ \hline
Baseline                          & 0.8685  & 0.9296    & 0.0279    & 0.0279    & 0.8685    & 0.9296    & 0.0198    & 0.0198    \\ \cline{1-9} 
Baseline w/ ALC                 & 0.7088       & 0.7935        & 0.0125        & 0.0125        & 0.7106        & 0.7792        & 0.0106        & 0.0106        \\ \cline{1-9} 
\Algnameabbr{}                   & 0.8008   & 0.8894    & 0.0084    & 0.0084    & 0.8327    & 0.9087    & 0.0095    & 0.0095    \\ \cline{1-9} 
\end{tabular}
\label{tab:exp_trial}
\end{table*}

\noindent\textbf{Results of patient-trial matching.} 
Table~\ref{tab:exp_trial} summarizes the performance of \Algnameabbr{} and the two baseline methods on the testing set. The prediction results of all three models are computed based on 100\% matching patient-criterion pairs as mentioned in Section~\ref{sec:prob_def}.
Since the patient-trial matching results are derived from patient-criterion matching, we can observe a similar trend as in the patient-criterion matching task.
Overall, \Algnameabbr{} shows promise in addressing fairness issues related to the two different sensitive attributes, \textit{race} and \textit{gender}, while causing a slight performance drop in patient-trial matching compared to the baseline model and the ALC debiasing method.

\subsection{Analysis of sensitive hyper-parameter $\lambda_{FC}$}
\label{sec:sensitive_hyper}
\begin{figure*}[ht]
\vspace{-0.2cm}
\centerline{\includegraphics[width=1.\textwidth]{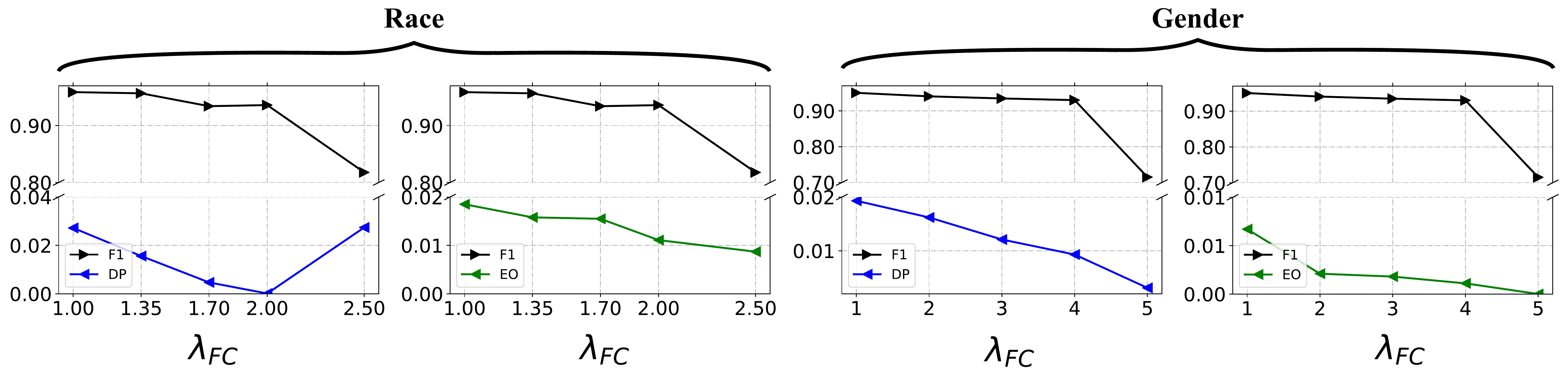}}
\vspace{-0.3cm}
\caption{Sensitive analysis of $\lambda_{FC}$ for \Algnameabbr{} on patient-criterion matching task with respect to the sensitive attributes of \textit{race} and \textit{gender}.}
\label{fig:seni_fc}
\end{figure*}

\vspace{-0.1cm}
In this section, we study the impact of the hyper-parameter $\lambda_{FC}$ in Eq.~\ref{eq:total_loss} to answer the research question \textbf{RQ2}. We conduct the sensitive analysis on patient-criterion matching since the results of patient-trial matching are derived from it. As shown in Figure~\ref{fig:seni_fc}, the value of $\lambda_{FC}$ does not significantly affect the patient-criterion matching performance up to a certain threshold, which is 2 for \textit{race} and 4 for \textit{gender}. After the certain value, the performance drop can be observed from Figure~\ref{fig:seni_fc}. Regarding the influence towards fairness, both the fairness metrics DP and EO can be improved when the value of $\lambda_{FC}$ increases, except for the DP fairness metric of the sensitive attribute \textit{race}. As the goal of \Algnameabbr{} is to mitigate the biased prediction outcomes against sensitive attributes while maintaining competitive performance in patient-criterion and patient-trial matching tasks, we can determine the appropriate value of $\lambda_{FC}$ by identifying the threshold before performance drop occurs.

\subsection{Case study: Fairness-related criteria}
\label{sec:case_study}
\vspace{-0.1cm}
To investigate the impact of the proposed \Algnameabbr{} framework on mitigating biased predictions, we compare the results between the baseline model and our \Algnameabbr{} for patients with different sensitive attributes $v_s$.
As shown in Table~\ref{tab:case_study}, the baseline model may provide biased predictions against the minority group for some eligibility criteria of clinical trials, which will exacerbate the underrepresentation problem.
For example, for the inclusion criterion in NCT03545607 ``Male or female subjects $\geq$ 18 years of age," the baseline model predicts a patient-criterion matching pair incorrectly when a female patient is older than 18 years old. This unfair outcome may be attributed to the biased knowledge encoded in the model, which associates the term ``Male" more strongly with the model's predictions. In contrast, our \Algnameabbr{} can predict the criterion for different sensitive groups fairly.

\begin{table*}[ht]
\small
\renewcommand{\arraystretch}{1.1}
\centering
\vspace{0.2 cm}
\caption{Case study: different patient-criterion matching results between the baselin model and \Algnameabbr{}.}
\vspace{-0.2 cm}
\begin{tabular}{c|c|c|c|c}
\hline
\textbf{Clinical trial}     & \textbf{Criterion}    & \textbf{Patient $v_s$}  & \textbf{Baseline}     & \textbf{\Algnameabbr{}}       \\
\midrule
\multirow{2}{*}{NCT03735979}     & \multirow{2}{*}{(I) Acute ischemic stroke patients.}   & Male      & \cmark    & \cmark    \\ 
                &               & Female    & \xmark    & \cmark    \\
\midrule
\multirow{2}{*}{NCT03545607}     & \multirow{2}{*}{(I) Male or female subjects $\geq$ 18 years of age.}   & Male      & \cmark    & \cmark    \\ 
                &               & Female    & \xmark    & \cmark    \\
\midrule
\multirow{2}{*}{NCT03876457}     & \multirow{2}{*}{(I) Eligible for thrombectomy or medical management.}   & White      & \xmark    & \cmark    \\ 
                &               & Others    & \cmark    & \cmark    \\
\midrule
\multirow{2}{*}{NCT03496883}     & \multirow{2}{*}{\makecell{(E) Patient suspected of not being able to comply with trial protocol \\ (e.g., due to alcoholism, drug dependency, or psychological disorder).}}   & White      & \xmark    & \cmark    \\ 
                &               & Others    & \cmark    & \cmark    \\
\bottomrule
\end{tabular}
\label{tab:case_study}
\end{table*}

\section{Conclusion and future work}
\vspace{-0.2cm}

We present \Algnameabbr{}, an innovative framework designed to tackle the issue of AI fairness in clinical trial matching with deep learning. Our approach includes a novel patient-criterion level fairness constraint that can help mitigate this problem. One of the unique features of our proposed framework is that it focuses on the discrepancy between inclusion and exclusion criteria for both the model predicting and alleviating unfair predictions. In doing so, it helps ensure that the patient selection process is unbiased, transparent, and equitable, leading to fairer and more reliable clinical trials outcomes. To demonstrate the effectiveness of \Algnameabbr{}, we conducted several experiments using real-world patient records with six stroke clinical trials. Our results indicated that the framework significantly improved two fairness metrics while only marginally affecting overall model performance. As part of our future research directions, we plan to explore various perspectives on bias mitigation in patient-trial matching, specifically the different forms of skew distribution in training data. We aim to expand our concept of fairness and explore how to mitigate bias that may arise due to distribution shift, a key challenge in machine learning applications. 


\makeatletter
\renewcommand{\@biblabel}[1]{\hfill #1.}
\makeatother

\bibliographystyle{vancouver}
\bibliography{amia}  

\end{document}